# Group-level Emotion Recognition using Transfer Learning from Face Identification

Alexandr Rassadin
National Research University Higher School of Economics
Laboratory of Algorithms and Technologies for Network Analysis, Nizhny Novgorod
Russia
arassadin@hse.ru

Alexey Gruzdev
National Research University Higher School of Economics
Nizhny Novgorod
Russia
alexey.s.gruzdev@gmail.com

Andrey Savchenko
National Research University Higher School of Economics
Laboratory of Algorithms and Technologies for Network Analysis, Nizhny Novgorod
Russia
avsavchenko@hse.ru

## ABSTRACT

In this paper we describe our algorithmic approach, which was used for submissions in the fifth Emotion Recognition in the Wild (EmotiW 2017) group-level emotion recognition sub-challenge. We extracted feature vectors of detected faces using the Convolutional Neural Network trained for face identification task, rather than traditional pre-training on emotion recognition problems. In the final pipeline an ensemble of Random Forest classifiers was learned to predict emotion score using available training set. In case when the faces have not been detected, one member of our ensemble extracts features from the whole image. During our experimental study, the proposed approach showed the lowest error rate when compared to other explored techniques. In particular, we achieved 75.4% accuracy on the validation data, which is 20% higher than the handcrafted feature-based baseline. The source code using Keras framework is publicly available.

## 1 INTRODUCTION

Image-based analysis of human activity and social behavior has attracted significant attention in the computer vision community during the past decade, since it can potentially produce a lot of interesting and useful applications, such as digital security surveillance, human computer interaction, street analytics, etc. A lot of above mentioned applications require to analyze emotions depending on particular use-cases. The main direction for this task is to identify faces in the photos and perform further detection of particular emotion [1].

There are various well-known methods for facial expression recognition, which contain different types of feature descriptors, such as facial Action Units (AUs) [2], geometric landmarks [3] or the Convolutional Neural Network (CNN) [4], [5]. However, only last years we see an increased interest to research 'Group' emotion in images [6], [7], [8]. Automatic happiness analysis of a group of people in an image using facial expression analysis is thoroughly discussed in [9]. This task can be helpful for crowd analytics, for security and social needs. During EmotiW 2016 challenge, several CNN-based approaches have been proposed to deal with inferring happiness intensity of a group of people in images [10], [11], [12].

This year, the challenge has advanced to extract group emotion from "in the wild" photos made in various environments from a broader valence range [13] including positive, neutral and negative emotions. In this work, we study various techniques for detection, feature extraction, emotion classification as well as end-to-end methods based on deep CNNs, which have proven themselves effectively in complicated computer vision tasks such as object detection, human pose estimation and image segmentation. We propose an ensemble of classifiers, which includes: 1) the transfer learning techniques to recognize facial emotion from rather different face identification task [12], [15], [16]; 2) processing of facial landmarks [3], [11]; and 3) extraction of the CNN bottleneck features from the whole photo.

The rest of the paper is organized as follows. Section 2 describes the proposed solution. In Section 3 the experimental results using the validation subset of this sub-challenge are discussed. Finally, in Section 4 we give the short summary and concluding remarks.

## 2 MATERIALS AND METHODS

### 2.1 Data overview

The main objective of EmotiW 2017 Group-level sub-challenge consists in accurate classification of a group's perceived emotion given an image, which represents natural scene with a group of people. The task is to assign each of the images one of 3 labels: Positive, Neutral or Negative.

The images in this sub-challenge are from the Group Affect Database 2.0 [13]. Statistics of the labels distribution over the train / validation subsets can be found in Table 1. Example of the training image showed in Fig. 1.

**Table 1:** EmotiW 2017 [2] data overview

| Subset | # frames | #Positive | #Negative | #Neutral |
|---|---|---|---|---|
| Train | 3630 | 1272 | 1159 | 1199 |
| Validation | 2065 | 773 | 564 | 728 |
| Test | 772 | | | |

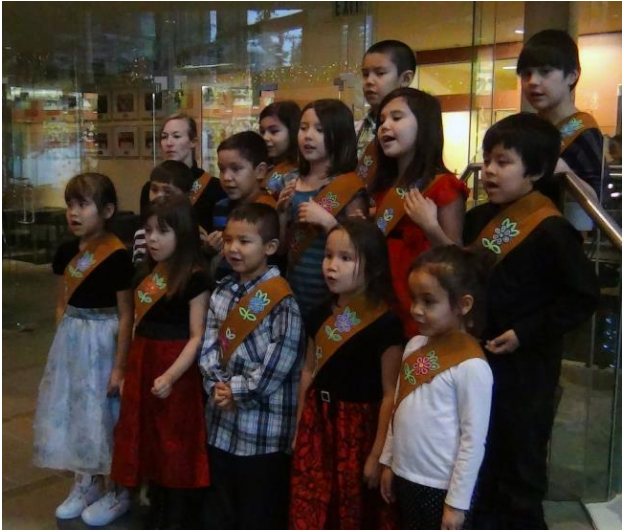

**Figure 1:** Example of validation image with neutral label.

One of the distinctive characteristic of this particular data is that images can be from positive social events such as convocations, marriages, party or neutral event such as meetings or negative events such as funeral, protests etc. The challenging property of the data is that images include faces of different scales, with illumination changes, object occlusions, various ages and non-trivially global context.

**2.2 Face Detection**

First block of our solution is a face detector. In particular, we need to find all faces on the group photo for the further facial expression analysis. At this stage, we decide to find faces with the traditional algorithms, namely, the Viola-Jones cascade classifiers based on the Haar features [17] from OpenCV library, and linear classifiers over an image pyramid in sliding window manner based on the Histogram of Oriented Gradients (HOG) descriptor [18] from DLIB library. In our experiments, we have found that on the most accurate face detector for EmotiW training and validation data is the DLIB frontal detector.

We tried recently open-sourced technique called TinyFace Detector [18] which is the deep ResNet-101 CNN [20] trained on WIDER Face dataset. In particular, this detector obtained much more correct facial regions when compared to the classical techniques. However, in our experimental study emotions on such tiny faces were poorly recognized. Hence, in the final pipeline we decided to use conventional approach. Namely, we extracted the faces using the detector from DLIB. However, sometimes it was not able to detect any of the faces from the training/validation/test subsets. In order to obtain sufficient amount of training data, we decided to use both DLIB and OpenCV face detection in sequential manner: first we run frontal face detector from DLIB library and if it does not return any face detections then we try to run OpenCV Haar cascades. This particular choice was done by practical observation that Haar cascade tends to return a lot of false positives, which can negatively affect the whole solution.

**2.3 Deep face feature extractor**

One of the most important parts of the presented algorithm is the deep face feature extractor. Nowadays the feature extraction is typically implemented with the transfer learning methods [21], [22]. The CNN is trained with an external large dataset, e.g., one of the previous EmotiW datasets [5] or FER2013 [23]. The outputs of the CNN's last layer for the input image and each reference image are used as their feature vectors. After that, conventional machine learning techniques, e.g., support vector machines or random forests are applied for these features to recognize facial expressions. In fact, the state-of-the-art performance on the EmotiW dataset has been achieved by such transfer learning approach [24].

However, in this paper we propose to use CNN trained to identify faces rather than recognize emotions. In fact, the face recognition task is much better studied when compared to emotion recognition. As a result, the volume of the external face identification datasets available to train a deep CNN is much higher. Since we analyse human faces, we assumed that the features for face recognition and facial expression analysis can be rather similar.

As a result, we decided to extract facial features using the VGGFace neural network [14], which was pre-trained for face recognition using the large VGG face dataset (2.6M images of 2622 identities). Each of the facial images was resized to 224x224 resolution, fed to the input of the VGGFace CNN. The 512 outputs at "avgpool" and 4096 outputs at "fc7" layers are stored in the corresponding feature vectors. Experimentally, we have observed that this CNN sometimes performs better if we performed channel permutation (RGB to BGR). It was decided to include this trick into the final pipeline, so we get two "avgpool" and two "fc7" vectors (in total, 4 feature sets) per each face.

After extracting CNN features for each detected face, we computed the final feature vector of the whole image as the median of corresponding features of individual frames. We also tested the more traditional averaging technique with computation of the mean feature vector, but it degrades the accuracy on 1-2% in all our experiments.

**2.4 Face landmarks estimation**

The next building block of our solution is the facial landmarks estimation. Facial landmarks are found with the help of DLIB implementation of the method described in the paper [25], which was trained on the iBUG 300-W face landmark dataset. As an example of feature engineering, we have found that beyond features extracted by CNN, it can be helpful to use distances between facial landmarks.

In particular, we computed all unique pairwise distances between extracted 68 facial landmarks and normalized these distances to prevent the dependence on the facial region size. To do this, we simply divide all distances by the maximum of the computed distances between found landmarks. During experiments, we also tried to normalize landmarks distances by mean distance value, however, such normalization caused small accuracy degradation.

### 2.5 Proposed approach

By using described features we learned the final classifier which predicts the group image emotion label. In our work we implemented most popular classifiers including Logistic Regression, Support Vector Regression, Gradient Boosting Trees and Random Forest. It was experimentally identified that the best choice of the trainable classifier is Random Forest, which performed the best on the validation data.

The final pipeline is shown in Fig. 2.

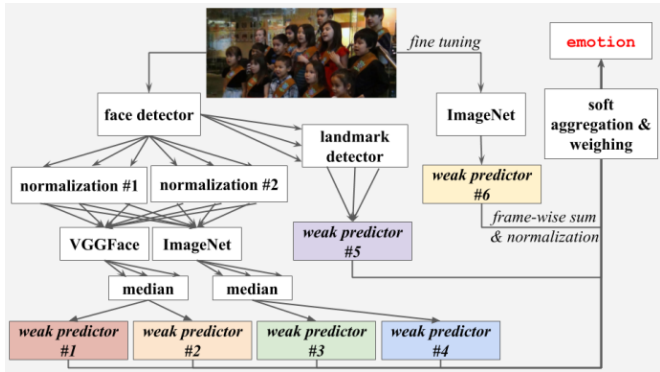

**Figure 2:** Proposed approach

It was constructed as an ensemble of four Random Forest classifiers trained on VGGFace "avgpool" (before the first fully connected layer) and "fc7" features for RGB and BGR representation of the face. An additional Random Forest classifier was trained on the normalized landmark distances. To tackle the problem of images where our detectors failed to extract any face, we fine-tuned a VGG16 model [26] pre-trained on ImageNet dataset [27] to perform the end-to-end classification for the full image. This classifier is also included in the ensemble. To obtain the final decision, the outputs of individual classifiers are weighted based on their accuracy estimated on the validation data. In addition, we tried to add the Random Forest classifier on top of scoring of individual predictors but it caused small degradation in accuracy. As such addition of a classifier only complicated our pipeline, we did not include the complex fusion of classifier confidences/scores into the final decision (Fig. 2).

## 3 RESULTS AND DISCUSSION

Classification accuracy is used as the metric in the challenge. All deep CNNs in this experiment were implemented using Keras library on top of the TensorFlow framework. In the experimental study the proposed approach (Fig. 2) was compared with the baseline support vector regression model for the CENTRIST scene descriptor [6]. For the CNN network in our pipeline (weak predictor #6) which classifies the whole group image, we used the Adagrad optimizer with learning rate of 1e-4, batch size of 40, aggressive data augmentation (random horizontal flip, ±10° of rotation, up to 10% zoom, channel shift by ±5) and trained this network for 30 epochs.

**Table 2:** Experimental results on the validation data

| Method | Accuracy |
| --- | --- |
| CENTRIST, SVM [6] (baseline) | 52.97% |
| VGG-16 (end-to-end) | 64.11% |
| VGG-19 (end-to-end) | 27.64% |
| VGGFace (end-to-end) | 65.42% |
| VGGFace features, SVM | 64.90 % |
| ResNet-50 (end-to-end) | 62.65% |
| XCeption (end-to-end) | 60.18% |
| VGGFace features, Logistic Regression | 67.24% |
| VGGFace features, SVM | 65.41% |
| VGGFace features, Multi-Layered Perceptron (1 hidden layer) | 68.09% |
| VGGFace features (BGR,avgpool), Gradient Boosting Tree | 69.53% |
| *VGGFace features (RGB, fc7), Random Forest* | *67.62%* |
| *VGGFace features (BGR, fc7), Random Forest* | *68.18%* |
| *VGGFace features (RGB, avgpool), Random Forest* | *69.78%* |
| *VGGFace features (BGR,avgpool), Random Forest* | *70.11%* |
| *Landmark features, Random Forest* | *65.16%* |
| *VGG-16 for the whole image* | *65.89%* |
| Proposed ensemble (except the VGG-16 for the whole image) | 72.77% |
| Proposed ensemble, Tiny faces | 66.51% |
| **Proposed ensemble, HOG/Viola-Jones faces** | **75.39%** |

In addition, we considered traditional approach with the end-to-end learning by fine-tuning several modern CNNs (VGG-16 and VGG-19 [26], VGGFace [14], ResNet-50 [20] and Xception [28]) using the detected faces. In this case it is assumed that all the faces detected in particular photo are characterized with the same emotion label as the whole image. The standard fine-tuning procedure was implemented: we removed the last (classification) layer, added a bottleneck fully-connected layer with 1024 neurons, dropout layer with probability 0.5 and final fully-connected classifier for 3-class classification problem. The following hyperparameters were chosen: learning rate 1e-4, Adam optimizer, batch size = 16. We trained such networks for 100 epochs with standard data augmentation: random crops, scale and horizontal flip with probability 0.5. The final decision is made by the simple voting of the classifier outputs for individual faces.

In Table 2 we report the comparative analysis of different methods which we explored during the EmotiW challenge. Here we marked by italic the individual classifiers, which are the members of the final ensemble.

Based on these results, we can draw the following conclusions. First, the end-to-end deep CNNs are better than the baseline, however they are worse than the classifiers in our ensemble. It can be explained by a wrong assumption that all faces in an image share the same emotion label as the whole group. Moreover, this approach is not designed to model scene context which has significant influence on the group emotion label. In the end-to-end training of VGG-19 we observed the typical overfitting case. As the more lightweight VGG-16 network was able to obtain higher accuracy (64.11%), we decided not to examine the VGG-19 architecture anymore.

Second, it should be noted that using the Tiny Face Detector [19] in ensemble of classifiers causes 9% less accuracy when compared to the final pipeline with traditional face detection techniques from DLIB and OpenCV libraries. We believe that the reason for such poor quality is the data which was used for training VGGFace [12]. The network was trained on rather large facial regions of the high quality (without occlusions, background cluttering, etc.) and is not appropriate for very small faces. Thus, the problem can be solved by replacing the pre-trained VGGFace feature extractor with another feature extractor, specially learned to identify tiny faces.

Thirdly, we examined that the recognition accuracy can be improved by adding new classifier into ensemble if its sole accuracy is rather high (not less than 65%). For example, adding the weak classifier (VGG-16 trained for the whole group image) made it possible to decrease error rate approximately 3%.

Finally, the proposed pipeline (Fig. 2) achieves **75.4%** accuracy on the validation set. The confusion matrix of our ensemble is presented in Fig. 3. One can note that our approach performs better on Negative and Positive classes rather than on Neutral one. It can be explained because the neutral group emotion is much more difficult to define. The same effect we observed while testing practically all other models. The only one exception is the processing of tiny faces [19], for which we obtained practically identical accuracy (66-67%) for each class.

## 4 CONCLUSIONS

In this paper we describe a solution used by our team for the group-level emotion recognition sub-challenge of EmotiW2017. Our algorithm (Fig. 2) includes detecting faces using classical Viola-Jones cascades and HOG features, detecting facial landmarks, extracting facial features using deep CNN trained for face identification task, computing the median of these features in order to deal with arbitrary number of faces in a group photo. The final decision is obtained by an ensemble of weighted Random Forest classifiers.

It is necessary to highlight that the training set of emotional images was used only to train the individual Random Forests. Hence, our method is not end-to-end, because the EmotiW 2017 dataset is not so large to learn competitive feature extractor. Our approach benefits from aggregating face representations without associating each of them with the label of the whole image. We have found that this particular approach can be useful for 'in the wild' group images, where it is not always true that each person shares the same emotion of a group.

Our pipeline achieved 75.4% classification accuracy on the validation data, which is practically 23% higher when compared to the baseline (CENTRIST feature detector and SVM [6]). Our model achieves 78.53% on the Test set (compare to 53.62% of the baseline with CENTRIST descriptors). Our Keras code is publicly available at https://github.com/arassadin/emotiw2017.

## ACKNOWLEDGMENTS

The article was prepared within the framework of the Academic Fund Program at the National Research University Higher School of Economics (HSE) in 2017 (grant No. 17-05-0007) and is supported by the Russian Academic Excellence Project "5–100" and Russian Federation President grant no. MD-306.2017.9.

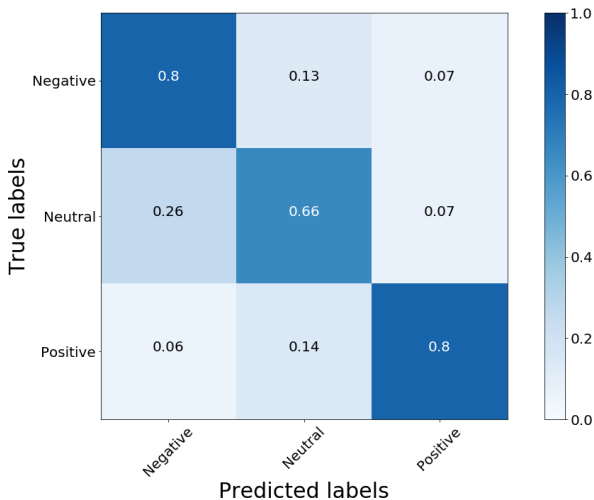

**Figure 3:** Confusion matrix of the final solution.

## REFERENCES
[1] E. Sariyanidi, H. Gunes, and A. Cavallaro. 2015. Automatic analysis of facial affect: A survey of registration, representation, and recognition. IEEE Transactions on Pattern Analysis and Machine Intelligence, 37(6), pp. 1113–1133
[2] S. Velusamy, H. Kannan, B. Anand, A. Sharma and B. Navathe. 2011. A method to infer emotions from facial action units. In *Proceedings of* IEEE *International Conference on Acoustics, Speech and Signal Processing (ICASSP)*, pp. 2028-2031
[3] Y. Hu, Z. Zeng, L. Yin, X. Wei, J. Tu and T. S. Huang. 2008. A study of non-frontal-view facial expressions recognition. In *Proceedings of 19th International Conference on Pattern Recognition (ICPR)*, pp. 1-4
[4] B. Kim, H. Lee, J. Roh and S. Lee, Hierarchical Committee of Deep CNNs with Exponentially-Weighted Decision Fusion for Static Facial Expression

Recognition. In *Proceedings of 18th ACM International Conference on Multimodal Interaction (ICMI)*, pp. 427-434

[5] G. Levi, and T. Hassner. 2015. Emotion recognition in the wild via convolutional neural networks and mapped binary patterns. In *Proceedings of 2015 ACM on International Conference on Multimodal Interaction (ICMI)*, pp. 503-510

[6] A. Dhall, J. Joshi, K. Sikka, R. Goecke and N. Sebe. 2015. The more the merrier: Analysing the affect of a group of people in images. In *Proceedings of 11th IEEE International Conference and Workshops on Automatic Face and Gesture Recognition (FG)*, vol. 1, pp. 1-8

[7] X. Huang, A. Dhall, X. Liu, G. Zhao, J. Shi, R. Goecke, M. Pietikäinen. 2016. Analyzing the affect of a group of people using multi-modal framework. arXiv preprint arXiv:1610.03640

[8] X. Huang, A. Dhall, G. Zhao, R. Goecke, M. Pietikäinen. 2015. Riesz-based Volume Local Binary Pattern and A Novel Group Expression Model for Group Happiness Intensity Analysis. In *Proceedings of British Machine Vision Conference*, pp. 34.1-34.13

[9] A. Dhall, R. Goecke and T. Gedeon. 2015. Automatic Group Happiness Intensity Analysis. In *IEEE Transactions on Affective Computing*, vol. 6, no. 1, pp. 13-26

[10] J. Li, S. Roy, J. Feng and T. Sim. 2016. Happiness level prediction with sequential inputs via multiple regressions. In *Proceedings of the 18th ACM International Conference on Multimodal Interaction (ICMI)*, pp. 487-493

[11] V. Vonikakis, Y. Yazici, V. D. Nguyen and S. Winkler. 2016. Group happiness assessment using geometric features and dataset balancing. In *Proceedings of the 18th ACM International Conference on Multimodal Interaction (ICMI)*, pp. 479-486

[12] B. Sun, Q.Wei, L. Li, Q. Xu, J. He and L.Yu. 2016. LSTM for dynamic emotion and group emotion recognition in the wild. In *Proceedings of the 18th ACM International Conference on Multimodal Interaction (ICMI)*, pp. 451-457

[13] A. Dhall, R. Goecke, S. Ghosh, J. Joshi, J. Hoey and T. Gedeon. 2017. From Individual to Group-level Emotion Recognition: EmotiW 5.0. ACM ICMI

[14] O. M. Parkhi, A. Vedaldi, A. Zisserman. 2015. Deep face recognition. In *Proceedings of the British Machine Vision Conference*

[15] A.V. Savchenko. 2016. *Search Techniques in Intelligent Classification Systems*. Springer, ISBN: 978-3-319-30515-8

[16] A.V. Savchenko. 2017. Maximum-likelihood approximate nearest neighbor method in real-time image recognition. *Pattern Recognition*, vol. 61, pp. 459-469

[17] P. Viola and M. Jones. 2001. Rapid object detection using a boosted cascade of simple features. In *Proceedings of the IEEE International Conference on Computer Vision and Pattern Recognition (CVPR)*, vol. 1

[18] N. Dalal and B. Triggs. 2005. Histograms of oriented gradients for human detection. In *Proceedings of the IEEE International Conference on Computer Vision and Pattern Recognition (CVPR)*, vol. 1, pp. 886-893

[19] P. Hu and D. Ramanan. 2017. Finding Tiny Faces. arXiv preprint arXiv:1612.04402

[20] K. He, X. Zhang, S. Ren and J. Sun. 2016. Deep residual learning for image recognition. In *Proceedings of the IEEE International Conference on Computer Vision and Pattern Recognition (CVPR)*, pp. 770-778

[21] P. Sermanet, D. Eigen, X. Zhang, M. Mathieu, R. Fergus and Y. LeCun. 2013. Overfeat: Integrated recognition, localization and detection using convolutional networks. arXiv preprint arXiv:1312.6229.

[22] A. V. Savchenko. 2017. Deep Convolutional Neural Networks and Maximum-Likelihood Principle in Approximate Nearest Neighbor Search. In *Proceedings of Iberian Conference on Pattern Recognition and Image Analysis*, L.A. Alexandre et al. (Eds.). Lecture Notes in Computer Science, vol. 10255. Springer, pp. 42–49.

[23] O. Arriaga and P.G. Ploger. 2017. Real-time Convolutional Neural Networks for Emotion and Gender Classification. In *Proceedings of Free and Open Source software Conference (FrOSCon)*

[24] H. Kaya, F. Gürpınar, A. A. Salah. 2017. Video-based emotion recognition in the wild using deep transfer learning and score fusion. *Image and Vision Computing*, doi: http://dx.doi.org/10.1016/j.imavis.2017.01.012

[25] V. Kazemi and J. Sullivan. 2014. One Millisecond Face Alignment with an Ensemble of Regression Trees. In *Proceedings of IEEE International Conference on Computer Vision and Pattern Recognition (CVPR)*, pp. 1867-1874

[26] K. Simonyan and A. Zisserman. 2015. Very deep convolutional networks for large-scale image recognition. arXiv preprint arXiv:1409.1556

[27] O. Russakovsky, J. Deng, H. Su, J. Krause, S. Satheesh, S. Ma, Z. Huang, A. Karpathy, A. Khosla, M. Bernstein, A. C. Berg, L. Fei-Fei. 2015. ImageNet Large Scale Visual Recognition Challenge. *International Journal of Computer Vision (IJCV)*, vol. 115, pp. 211-252, doi: 10.1007/s11263-015-0816-y

[28] F. Chollet. 2016. Xception: Deep Learning with Depthwise Separable Convolutions. arXiv preprint arXiv:1610.02357.